\def\year{2018}\relax
\documentclass[letterpaper]{article} %DO NOT CHANGE THIS
\usepackage{aaai18}  %Required
\usepackage{times}  %Required
\usepackage{helvet}  %Required
\usepackage{courier}  %Required
\usepackage{url}  %Required
\usepackage{graphicx}  %Required
\usepackage{booktabs}       % professional-quality tables
\usepackage{amsmath}  
\usepackage{amsfonts}       % blackboard math symbols
\usepackage{subfig}

\newcommand{\eg}{\mbox{e.g.} }

\newcommand{\D}{\mathcal{D}}
\newcommand{\N}{\mathcal{N}}
\newcommand{\E}{\mathbb{E}}
\newcommand{\bs}{\boldsymbol}
\newcommand{\bx}{\textbf{x}}

\newcommand{\by}{\textbf{y}}

\newcommand{\bz}{\textbf{z}}

\begin{document}
\title{Deep Learning from Crowds}
\author{Filipe Rodrigues, Francisco C. Pereira\\
Dept. of Management Engineering, Technical University of Denmark\\
Bygning 116B, 2800 Kgs. Lyngby, Denmark\\
rodr@dtu.dk, camara@dtu.dk\\
}
\maketitle
\begin{abstract}
Over the last few years, deep learning has revolutionized the field of machine learning by dramatically improving the state-of-the-art in various domains. However, as the size of supervised artificial neural networks grows, typically so does the need for larger labeled datasets. Recently, crowdsourcing has established itself as an efficient and cost-effective solution for labeling large sets of data in a scalable manner, but it often requires aggregating labels from multiple noisy contributors with different levels of expertise. In this paper, we address the problem of learning deep neural networks from crowds. We begin by describing an EM algorithm for jointly learning the parameters of the network and the reliabilities of the annotators. Then, a novel general-purpose crowd layer is proposed, which allows us to train deep neural networks end-to-end, directly from the noisy labels of multiple annotators, using only backpropagation. We empirically show that the proposed approach is able to internally capture the reliability and biases of different annotators and achieve new state-of-the-art results for various crowdsourced datasets across different settings, namely classification, regression and sequence labeling.
\end{abstract}

\section{Introduction}

In the last decade, deep learning has made major advances in solving artificial intelligence problems in different domains such as speech recognition, visual object recognition, object detection and machine translation \cite{schmidhuber2015deep}. This success is often attributed to its ability to discover intricate structures in high-dimensional data \cite{lecun2015deep}, thereby making it particularly well suited for tackling complex tasks that are often regarded as characteristic of humans, such as vision, speech and natural language understanding. However, typically, a key requirement for learning deep representations of complex high-dimensional data is large sets of labeled data. Unfortunately, in many situations this data is not readily available, and humans are required to manually label large collections of data. 

On the other hand, in recent years, crowdsourcing has established itself as a reliable solution to annotate large collections of data. Indeed, crowdsourcing platforms like Amazon Mechanical Turk\footnote{http://www.mturk.com} and Crowdflower\footnote{http://crowdflower.com} have proven to be an efficient and cost-effective way for obtaining labeled data \cite{Snow2008,buhrmester2011amazon}, especially for the kind of human-like tasks, such as vision, speech and natural language understanding, for which deep learning methods have been shown to excel. Even in fields like medical imaging, crowdsourcing is being used to collect the large sets of labeled data that modern data-savvy deep learning methods enjoy \cite{greenspan2016guest,albarqouni2016aggnet,guan2017said}. 
However, while crowdsourcing is scalable enough to allow labeling datasets that would otherwise be impractical for a single annotator to handle, it is well known that the noise associated with the labels provided by the various annotators can compromise practical applications that make use of such type of data \cite{Sheng2008,DonmezCarbonell2008}. Thus, it is not surprising that a large body of the recent machine learning literature is dedicated to mitigating the effects of the noise and biases inherent to such heterogeneous sources of data (e.g. \citeauthor{Yan2014} \shortcite{Yan2014}; \citeauthor{albarqouni2016aggnet} \shortcite{albarqouni2016aggnet}; \citeauthor{guan2017said} \shortcite{guan2017said}). 

When learning deep neural networks from the labels of multiple annotators, typical approaches rely on some sort of label aggregation mechanisms prior to training. In classification settings, the simplest and most common approach is to use majority voting, which naively assumes that all annotators are equally reliable. More advanced approaches, such as the one proposed in \cite{DawidSkeene1979} and other variants (e.g. \citeauthor{ipeirotis2010quality} \shortcite{ipeirotis2010quality}; \citeauthor{whitehill2009whose} \shortcite{whitehill2009whose}) jointly model the unknown biases of the annotators and their answers as noisy versions of some latent ground truth. Despite their improved ground truth estimates over majority voting, recent works have shown that jointly learning the classifier model and the annotators noise model using EM-style algorithms generally leads to improved results \cite{Raykar2010,albarqouni2016aggnet}. 

In this paper, we begin by describing an EM algorithm for learning deep neural networks from crowds in multi-class classification settings, highlighting its limitations. Then, a novel \textit{crowd layer} is proposed, which allows us to train neural networks end-to-end, directly from the noisy labels of multiple annotators, using only backpropagation. This alternative approach not only allows us to avoid the additional computational overhead of EM, but also leads to a general-purpose framework that generalizes trivially beyond classification settings. Empirically, the proposed crowd layer is shown to be able to automatically distinguish the good from the unreliable annotators and capture their individual biases, thus achieving new state-of-the-art results in real data from Amazon Mechanical Turk for image classification, text regression and named entity recognition. As our experiments show, when compared to the more complex EM-based approaches and other approaches from the state of the art, the crowd layer is able to achieve comparable or, in many cases, significantly superior results.  

\section{Related work}

The increasing popularity of crowdsourcing as a way to label large collections of data in an inexpensive and scalable manner has led to much interest of the machine learning community in developing methods to address the noise and trustworthiness issues associated with it. In this direction, one of the key early contributions is the work of Dawid and Skene \shortcite{DawidSkeene1979}, who proposed an EM algorithm to obtain point estimates of the error rates of patients given repeated but conflicting responses to medical questions. This work was the basis for many other variants for aggregating labels from multiple annotators with different levels of expertise, such as the one proposed in \cite{whitehill2009whose}, which further extends Dawid and Skene's model by also accounting for item difficulty in the context of image classification. Similarly, Ipeirotis et al. \shortcite{ipeirotis2010quality} propose using Dawid and Skene's approach to extract a single quality score for each worker that allows to prune low-quality workers. The approach proposed in our paper contrast with this line of work, by allowing neural networks to be trained directly on the noisy labels of multiple annotators, thereby avoiding the need to resort to prior label aggregation schemes. 

Despite the generality of label aggregation approaches described above, which can be used in combination with any type of machine learning algorithm, they are sub-optimal when compared to approaches that also jointly learn the classifier itself. One of the most prominent works in this direction is the one of Raykar et al. \shortcite{Raykar2010}, who proposed an EM algorithm for jointly learning the levels of expertise of different annotators and the parameters of a logistic regression classifier, by modeling the ground truth labels as latent variables. This idea was later extended to other types of models such as Gaussian process classifiers \cite{Rodrigues2014}, supervised latent Dirichlet allocation \cite{rodrigues2017learning} and, recently, to convolutional neural networks with softmax outputs \cite{albarqouni2016aggnet}. In this paper, we begin by describing a generalization of the approach in \cite{albarqouni2016aggnet} to multi-class settings, highlighting some of the technical difficulties associated with it. Then, a novel type of neural network layer is proposed, which allows the training of deep neural networks directly from the noisy labels of multiple annotators using pure backpropagation. This contrasts with most of works in the literature, which rely on more complex iterative procedures based on EM. Furthermore, the simplicity of the proposed approach allows for straightforward extensions to regression and structured prediction problems. 

Recently, \citeauthor{guan2017said} \shortcite{guan2017said} also proposed an approach for training deep neural networks that exploits information about the annotators. The idea is to model the multiple experts individually in the neural network and then, while keeping their predictions fixed, independently learning averaging weights for combining them using backpropagation. Like our proposed approach, this two-stage procedure does not require an EM algorithm to estimate the annotators weights. However, while our approach has the ability to capture the biases of the different annotators (\eg confusing class 2 with class 4) and correct them, the approach in \cite{guan2017said} only learns how to combine the predicted answers of multiple annotators by weighting them differently. Moreover, its two-stage learning procedure increases the computation complexity of training, whereas in our proposed approach is kept the same. Lastly, while the work in \cite{guan2017said} focuses only on classification, we consider regression and structured prediction problems as well. 

Regarding applications areas for multiple-annotator learning, some of the most popular ones are: image classification \cite{Smyth1995,Welinder2010}, computer-aided diagnosis/radiology \cite{Raykar2010,greenspan2016guest}, object detection \cite{su2012crowdsourcing}, text classification \cite{rodrigues2017learning}, natural language processing \cite{Snow2008} and speech-related tasks \cite{parent2011speaking}. In this paper, we will use data from some of these areas to evaluate different approaches. %Given that these areas have seen dramatic improvements due to recent advances in deep learning \cite{lecun2015deep,schmidhuber2015deep}, developing novel efficient algorithms for learning deep neural networks from crowds is of great importance to the field. 
Given that these are precisely some of the areas that have seen the most dramatic improvements due to recent contributions in deep learning \cite{lecun2015deep,schmidhuber2015deep}, developing novel efficient algorithms for learning deep neural networks from crowds is of great importance to the field. 

\section{EM algorithm for deep learning from crowds}
\label{sec:em_algorithm}

Let $\D = \{\bx_n, \by_n\}_{n=1}^N$ be a dataset of size $N$, where for each input vector $\bx_n \in \mathbb{R}^D$ we are given a vector of crowdsourced labels $\by_n = \{y_n^r\}_{r=1}^R$, with $y_n^r$ representing the label provided by the $r^{th}$ annotator in a set of $R$ annotators. %Our goal is then to model the conditional distribution $p(\by_n | \bx_n)$. 
Following the ideas in \cite{Raykar2010,Yan2014}, we shall assume the existence of a latent true class $z_n$ whose value is, in this particular case, determined by a softmax output layer of a deep neural network parameterized by $\bs\Theta$, and that each annotator then provides a noisy version of $z_n$ according to $p(y_n^r|z_n, \bs\Pi^r) = Multinomial(y_n^r|\bs\pi_{z_n}^r)$. This formulation corresponds to keeping a per-annotator confusion matrix $\bs\Pi^r = (\bs\pi_1^r,\dots,\bs\pi_C^r)$ to model their expertise, where $C$ denotes the number of classes. Further assuming that annotators provide labels independently of each other, we can write the complete-data likelihood as
\begin{align}
p(\D,\bz|\bs\Theta,\{\bs\Pi^r\}_{r=1}^R) &= \prod_{n=1}^N p(z_n | \bx_n, \bs\Theta) \prod_{r=1}^R p(y_n^r|z_n, \bs\Pi^r).\nonumber
\end{align}

Based on this formulation, we can derive an Expectation-Maximization (EM) algorithm for jointly learning the reliabilities of the annotators $\bs\Pi^r$ and the parameters of the neural network $\bs\Theta$. The expected-value of the complete-data log-likelihood under a current estimate of the posterior distribution over latent variables $q(z_n)$ is given by
\begin{align}
&\E\Big[\ln p(\D,\bz|\bs\Theta,\bs\Pi^1,\dots,\bs\Pi^R)\Big] = \nonumber\\
& \sum_{n=1}^N \sum_{z_n} q(z_n) \ln \bigg( p(z_n | \bx_n, \bs\Theta) \prod_{r=1}^R p(y_n^r|z_n, \bs\Pi^r) \bigg),
\label{eq:expected_loglikelihood}
\end{align}
where the posterior $q(z_n)$ is obtained by making use of Bayes' theorem given the previous estimate of the model parameters $\{\bs\Theta_{\mbox{\tiny old}},\bs\Pi_{\mbox{\tiny old}}^1,\dots,\bs\Pi_{\mbox{\tiny old}}^R\}$, yielding
\begin{align}
q(z_n = c) &\propto p(z_n = c | \bx_n, \bs\Theta_{\mbox{\tiny old}}) \prod_{r=1}^R p(y_n^r|z_n = c, \bs\Pi_{\mbox{\tiny old}}^r).\nonumber
\end{align}
This corresponds to the E-step of EM. In the M-step, we find the new maximum likelihood for the model parameters. The update for the annotators' reliability parameters is given by
\begin{align}
\pi_{c,l}^r = \frac{\sum_{n=1}^N q(z_n = c) \, \mathbb{I}(y_n^r = l)}{\sum_{n=1}^N q(z_n = c)},\nonumber
%\pi_{c,l}^r \propto \sum_{n=1}^N q(z_n = c) \, \mathbb{I}(y_n^r = l),\nonumber
\end{align}
where $\mathbb{I}(y_n^r = l)$ is an indicator function that takes the value 1 when $y_n^r = l$, and zero otherwise. In practice, since crowd annotators typically only label a small portion of the data, it is particularly important to carefully impose Dirichlet priors on each $\bs\pi_c^r$ and compute MAP estimates instead, in order to avoid numerical issues. 

As for estimating the parameters of the deep neural network $\bs\Theta$, we follow the approach in \cite{albarqouni2016aggnet} and use the noise-adjusted ground-truth estimates $q(z_n)$ to backpropagate the error through the network using standard stochastic optimization techniques such as stochastic gradient descent (SGD) or Adam \cite{kingma2014adam}. Kindly notice how this raises the important question of how to schedule the EM steps. If we perform one EM iteration per mini-batch, we risk not having enough evidence to estimate the annotators reliabilities. On the other hand, if we run SGD or Adam until convergence, then the computational overhead of EM becomes very large. In practice, we found that, typically, one EM iteration per training epoch provides good computational efficiency without compromising accuracy. However, this seems to vary among different datasets, thus making it hard to tune in practice.

One key fundamental aspect for the development of this EM approach was the probabilistic interpretation of the softmax output layer of deep neural networks for classification. Unfortunately, such probabilistic interpretation is typically not available when considering, for example, continuous output variables, thereby making it more difficult to generalize this approach to regression problems. Furthermore, notice that if the target variable is a sequence (or any other structured prediction output), then the marginalization over the latent variables in (\ref{eq:expected_loglikelihood}) quickly become intractable, as the number of possible label sequences to sum over grows exponentially with the length of the sequence. %In the follow section, a new alternative approach is proposed, which does not suffer from these issues. 

\section{Crowd layer}

\begin{figure}[ht!]
\centering
%\fbox{\rule[-.5cm]{0cm}{4cm} \rule[-.5cm]{4cm}{0cm}}
%\includegraphics[width=0.6\linewidth]{CrowdsLayers.png}
\includegraphics[width=1.0\linewidth]{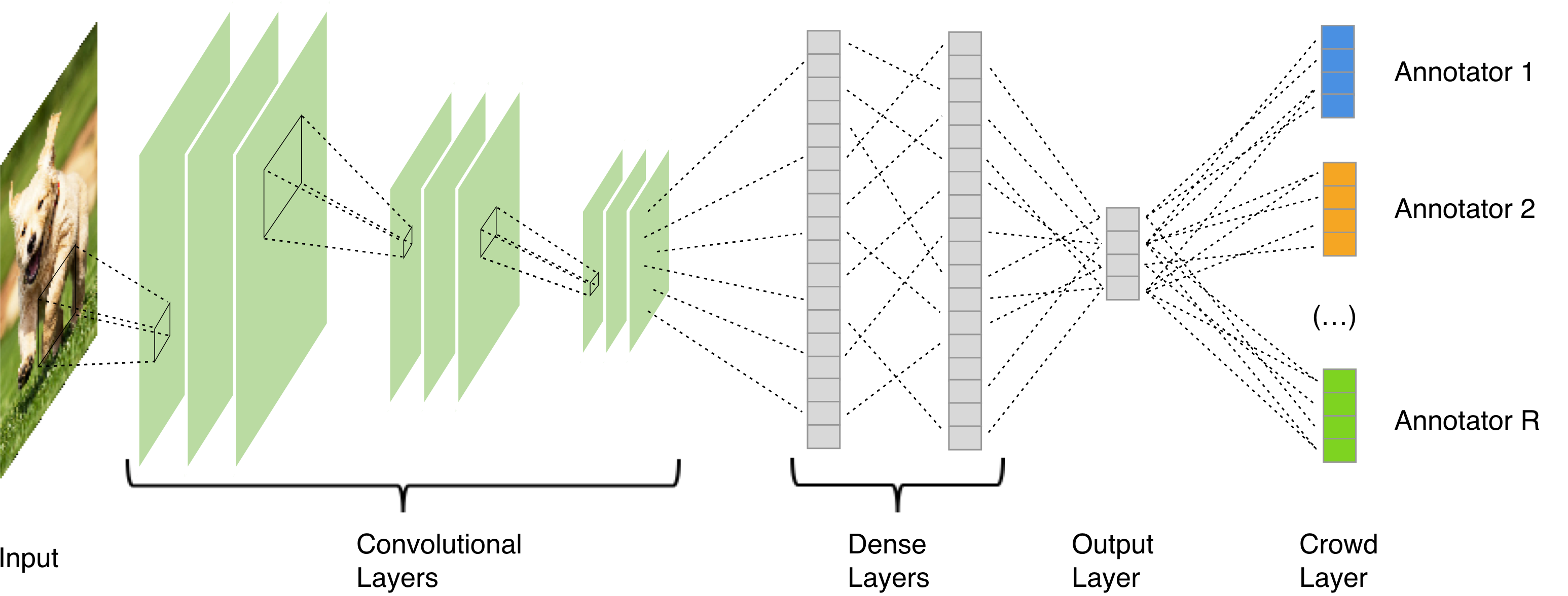}
\caption{Bottleneck structure for a CNN for classification with 4 classes and R annotators.}
\label{fig:network_example}
\end{figure}

In this section, we propose the \textit{crowd layer}: a special type of network layer that allows us to train deep neural networks directly from the noisy labels of multiple annotators, thereby avoiding some of the aforementioned limitations of EM-based approaches for learning from crowds. The intuition is rather simple. The crowd layer takes as input what would normally be the output layer of a deep neural network (e.g. softmax for classification, or linear for regression), and learns an annotator-specific mapping from the output layer to the labels of the different annotators in the crowd that captures the annotator reliabilities and biases. In this way, the former output layer becomes a bottleneck layer that is shared among the different annotators. Figure~\ref{fig:network_example} illustrates this bottleneck structure in the context of a simple convolutional neural network for classification problems with 4 classes and R annotators. 

The idea is then that when using the labels of a given annotator to propagate errors through the whole neural network, the crowd layer adjusts the gradients coming from the labels of that annotator according to his/her reliability by scaling them and adjusting their bias. In doing so, the bottleneck layer of the network now receives adjusted gradients from the different annotators' labels, which it aggregates and backpropagates further through the rest of the network. As it turns out, through this crowd layer, the network is able to account for unreliable annotators and even correct systematic biases in their labeling. Moreover, all of that can be done naturally within the  backpropagation framework. 

Formally, let $\bs\sigma$ be the output of a deep neural network with an arbitrary structure. Without loss of generality, we shall assume the vector $\bs\sigma$ to correspond to the output of a softmax layer, such that $\sigma_c$ corresponds to the probability of the input instance belonging to class $c$. The activation of the crowd layer for each annotator $r$ is then defined as $\textbf{a}^r = f_r(\bs\sigma)$, where $f_r$ is an annotator-specific function, and the output of the crowd layer simply as the softmax of the activations $o_c^r = e^{a_c^r} / \sum_{l=1}^C e^{a_l^r}$. 

The question is then how to define the function mapping $f_r(\bs\sigma)$. In the experiments section, we study different alternatives. For classification problems a reasonable assumption is to consider a matrix transformation, such that $f_r(\bs\sigma) = \textbf{W}^r \bs\sigma$, where $\textbf{W}^r$ is an annotator-specific matrix. Given a cost function $E(\textbf{o}^r, y^r)$ between the expected output of the $r^{th}$ annotator and its actual label $y^r$, we can compute the gradients $\partial E / \partial \textbf{a}^r$ at the activation $\textbf{a}^r$ for each annotator and backpropagate them to the bottleneck layer, leading to
\begin{align}
\frac{\partial E}{\partial \bs\sigma} = \sum_{r=1}^R \textbf{W}^r \frac{\partial E}{\partial \textbf{a}^r}. \nonumber
\end{align}
The gradient vector at the bottleneck layer then naturally becomes a weighted sum of gradients according to the labels of the different annotators. Moreover, if the annotator is likely to mislabel class $c$ as class $l$ (annotation bias), then the matrix $\textbf{W}^r$ can actually adjust the gradients accordingly. The problem of missing labels from some of the annotators can be easily addressed by setting their gradient contributions to zero. As for estimating the annotator weights $\{\textbf{W}^r\}_{r=1}^R$, since they parameterize the mapping from the output of the bottleneck layer $\bs\sigma$ to the annotators labels $\{\textbf{o}^r\}_{r=1}^R$, they can be estimated using standard stochastic optimization techniques such as SGD or Adam \cite{kingma2014adam}. Once the network is trained, the crowd layer can be removed, thus exposing the output of bottleneck layer $\bs\sigma$, which can readily be used to make predictions for unseen instances. 

An obvious concern with the approach described above is identifiability. Therefore, it is important to not over-parametrize $f_r(\bs\sigma)$, since adding parameters beyond necessary can make the output of the bottleneck layer $\bs\sigma$ lose its interpretability as a shared estimated ground truth. Another important aspect is parameter initialization. In our experiments, we found that the best practice is to initialize the crowd layer with identities, \mbox{i.e.} zeros for additive parameters, ones for scalar parameters, identity matrix for multiplicative matrices, etc. An alternative solution is to use regularization to force the parameters of the crowd layer to be close to identities. However, in some cases this might be an undesirable property. For example, if we consider a very biased annotator, then we do not wish to force the matrix $\textbf{W}^r$ to be close to the identity matrix. Based on our experiments, the initialization alternative provides the best results. Lastly, it should be noted that, as with EM-based approaches, %\cite{Raykar2010,albarqouni2016aggnet}
there is an implicit assumption that random or adversarial annotators do not constitute a vast majority (which generally holds in practice), %in which case the crowd layer would not be able to identify ``good" annotators, and it would eventually do no better than a random predictor. 
in which case the crowd layer would not perform better than a random predictor. 

A particularly important aspect to note, is that the framework described above is quite general. For example, it can be straightforwardly applied to sequence labeling problems without further changes, or be adapted to regression problems by considering univariate scalar and bias parameters per annotator in the crowd layer. 

\section{Experiments}
\label{sec:experiments}

The proposed crowd layer (CL) was implemented as a new type of layer in Keras \cite{chollet2015keras}, so that using it in practice requires only a single line of code. %Similarly, the EM-based approach described earlier and the approach from  were also implemented in Keras. 
Source code, datasets and demos for all experiments are provided at: \url{http://www.fprodrigues.com/}. %\footnote{Source code, datasets and demos available here: (see supplementary material)} 
%We evaluate the proposed crowd layer in 3 tasks: image classification, text regression and named entity recognition. 

\subsection{Image classification}

We begin by evaluating the proposed crowd layer in a more controlled setting, by using simulated annotators with different levels of expertise on a large image classification dataset consisting of 25000 images of dogs and cats from \cite{dogsvscatskaggle}, where the goal is to distinguish between the two species. Let the dog and cat classes be represented by 1 and 0, respectively. Since this is a binary classification task, we can easily simulate annotators with different levels of expertise by assigning them individual sensitivities $\alpha^r$ and specificities $\beta^r$, and sampling their answers from a Bernoulli distribution with parameter $\alpha^r$ if the true label is 1, and from a Bernoulli distribution with parameter $\beta^r$ otherwise. Using this procedure, we simulated a challenging scenario with 5 annotators with the following values of $\alpha^r = [0.6, 0.9, 0.5, 0.9, 0.9]$ and $\beta^r = [0.7, 0.8, 0.5, 0.2, 0.9]$. %When compared to the ground truth labels, the 5 simulated annotators revealed accuracies of 64.8\%, 84.7\%, 49.8\%, 54.7\% and 89.9\%. 

For this particular problem we used a fairly standard CNN architecture with 4 convolutional layers with 3x3 patches, 2x2 max pooling and ReLU activations. The output of the convolutional layers is then fed to a fully-connected (FC) layer with 128 ReLU units and finally goes to an output layer with a softmax activation. We use batch normalization \cite{ioffe2015batch} and apply 50\% dropout between the FC and output layers. The proposed approach further adds a crowd layer on top of the softmax output layer during training. The base architecture was selected from a set of possible configurations using the true labels by optimizing the accuracy on a validation set (consisting of 20\% of the train set) through random search. It is important to note that it is supposed to be representative of a set of typical approaches for image classification rather than being the single best possible architecture in the literature for this particular dataset. Furthermore, our main interest in this paper is the contribution of the crowd layer to the training of the neural network.

The proposed CNN with a crowd layer (referred to as ``DL-CL") is compared with: the multi-annotator approach from \cite{rodrigues2017learning} based on supervised latent Dirichlet allocation - ``MA-sLDA"; a CNN trained on the result of (hard) majority voting - ``DL-MV"; a CNN trained on the output of the label aggregation approach proposed by Dawid and Skene \shortcite{DawidSkeene1979} - ``DL-DS"; a CNN using the EM approach described earlier - ``DL-EM"; a CNN using the ``Doctor Net" approach from \cite{guan2017said} - ``DL-DN", which consists on training a CNN to predict the labels of the multiple annotators and then combining their predictions using majority voting; and, lastly, a CNN using the ``Weighted Doctor Net" approach from \cite{guan2017said} - ``DL-WDN", which is the best performing variant according to the original paper. This approach is similar to ``DL-DN" but additionally learns how to weight the predictions of the different annotators. Kindly see \cite{guan2017said} for further details.
%\begin{itemize}
%\item a CNN trained on the result of (hard) majority voting - ``DL-MV";
%\item a CNN trained on the output of the label aggregation approach proposed by Dawid and Skene \shortcite{DawidSkeene1979} - ``DL-DS";
%\item a CNN using the EM approach described earlier - ``DL-EM";
%\item a CNN using the ``Doctor Net" approach from \cite{guan2017said} - (``DL-DN"), which consists on training a CNN to predict the labels of the multiple annotators and then combining their predictions using majority voting;
%\item a CNN using the ``Weighted Doctor Net" approach from \cite{guan2017said} - (``DL-WDN"), which is the best performing version according to the original paper. This approach is similar to ``DL-DN" but additionally learns how to weight the predictions of the different annotators. Kindly see \cite{guan2017said} for further details.
%\end{itemize} 
As a reference point, we also compare with a CNN trained on true labels - ``DL-TRUE". 
We consider 3 variants of the proposed crowd layer (CL) with different annotator-specific functions $f_r$ with increasing number of parameters: a vector of per-class weights $\textbf{w}^r$, such that $f_r(\bs\sigma) = \textbf{w}^r \odot \bs\sigma$ (referred to as ``VW"); a vector of per-class biases $\textbf{b}^r$, such that $f_r(\bs\sigma) = \bs\sigma + \textbf{b}^r$ (``VB"); and a version with a matrix of weights $\textbf{W}^r$, such that $f_r(\bs\sigma) = \textbf{W}^r \bs\sigma$ (``MW").
%\begin{itemize}
%\item A vector of with per-class weights $\textbf{w}^r$ (``VW"): $f_r(\bs\sigma) = \textbf{w}^r \odot \bs\sigma$;
%%\item A vector of with per-class biases $\textbf{b}^r$ (``VB"): $f_r(\bs\sigma) = \bs\sigma + \textbf{b}^r$;
%\item A vector of with per-class weights $\textbf{w}^r$ and biases $\textbf{b}^r$ (``VW+B"): $f_r(\bs\sigma) = \textbf{w}^r \odot \bs\sigma + \textbf{b}^r$;
%\item A matrix of weights $\textbf{W}^r$ (``MW"): $f_r(\bs\sigma) = \textbf{W}^r \bs\sigma$.
%\end{itemize} 
In our experiments, we found that for approaches with more parameters than MW, such as $f_r(\bs\sigma) = \textbf{W}^r \bs\sigma + \textbf{b}^r$, identifiability issues start to occur. 

\begin{table*}[!htbp]
\caption{Accuracy results for classification datasets: Dogs vs. Cats and LabelMe.}
\label{table:classification_results}
\centering
\begin{tabular}{lll}
\toprule
Method & Dogs vs. Cats & LabelMe (MTurk)\\
\midrule
%sLDA-MV \cite{rodrigues2017learning} & - & 70.100 ($\pm$ 0.413) \\
MA-sLDAc \cite{rodrigues2017learning} & - & 78.120 ($\pm$ 0.397) \\
DL-MV & 71.377 ($\pm$ 1.123) & 76.744 ($\pm$ 1.208)  \\
DL-DS \cite{DawidSkeene1979} & 76.750 ($\pm$ 1.282) & 80.792 ($\pm$ 1.066)  \\
DL-EM \cite{albarqouni2016aggnet} & 80.184 ($\pm$ 1.454) & 82.677 ($\pm$ 0.981)  \\
DL-DN \cite{guan2017said} & 79.005 ($\pm$ 1.347) & 81.888 ($\pm$ 1.114)  \\
DL-WDN \cite{guan2017said} & 76.822 ($\pm$ 2.838) & 82.410 ($\pm$ 0.783)  \\
\midrule
DL-CL (VW) & 79.534 ($\pm$ 1.064) & 81.051 ($\pm$ 0.899)  \\
DL-CL (VW+B) & 79.688 ($\pm$ 1.406) & 81.886 ($\pm$ 0.893)  \\
DL-CL (MW) & 80.265 ($\pm$ 1.230) & 83.151 ($\pm$ 0.877)  \\
\midrule
DL-TRUE & 84.912 ($\pm$ 1.248) & 90.038 ($\pm$ 0.652)  \\
\bottomrule
\end{tabular}
\end{table*}

\begin{figure}[!htbp]
\centering
\subfloat[Sensitivities]{\includegraphics[width=0.47\linewidth]{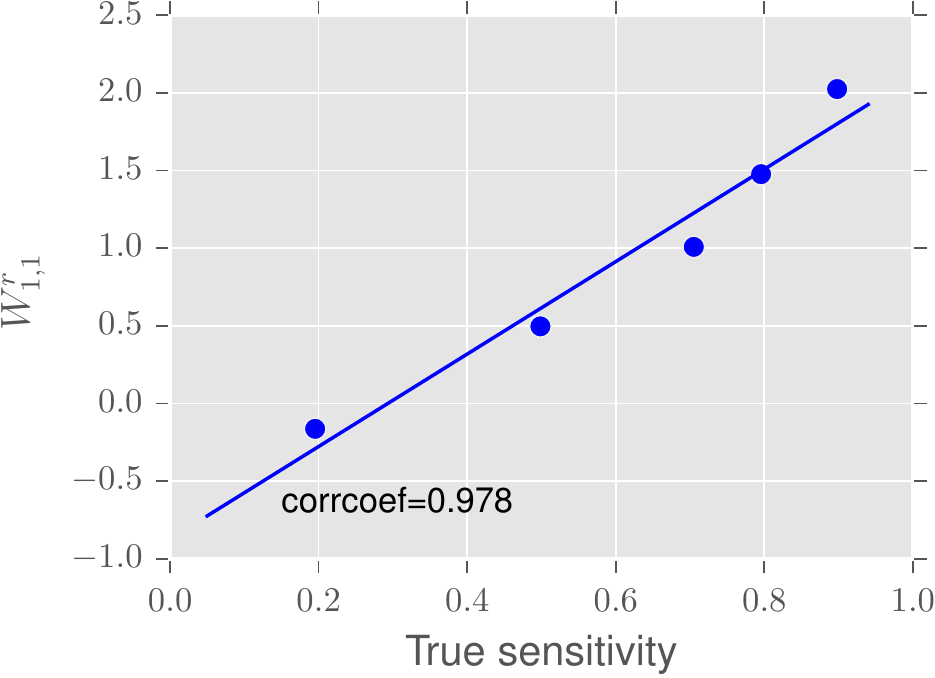}}\hspace{0.0cm}
\subfloat[Specificities]{\includegraphics[width=0.47\linewidth]{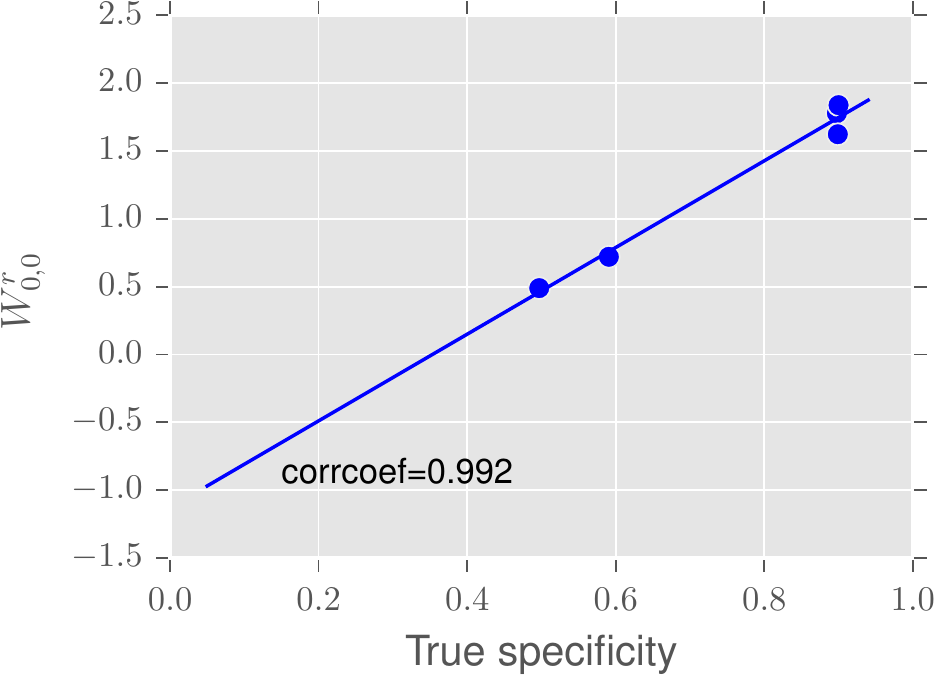}}
\caption{Comparison between the true sensitivities and specificities of the annotators and the diagonal elements of their weight matrices $\textbf{W}^r$ for the Dogs vs. Cats dataset.}
\label{fig:sens_spec_plot}
\end{figure}

\begin{figure*}[!htbp]
\centering
\subfloat[annot. 1]{\includegraphics[width=0.12\linewidth]{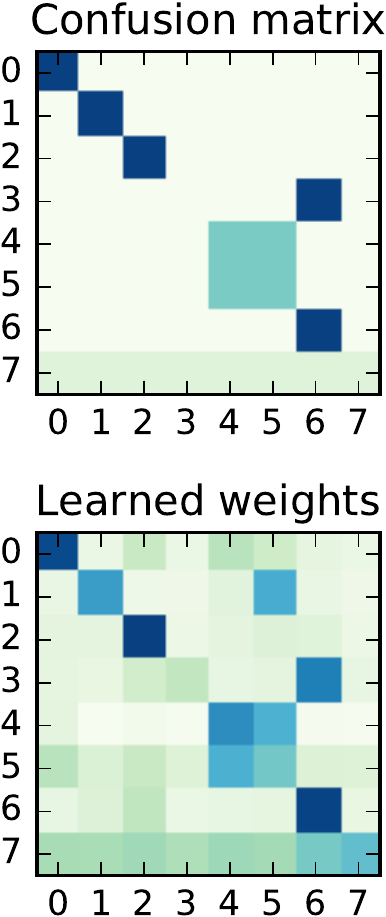}}\hspace{0.28cm}
\subfloat[annot. 9]{\includegraphics[width=0.12\linewidth]{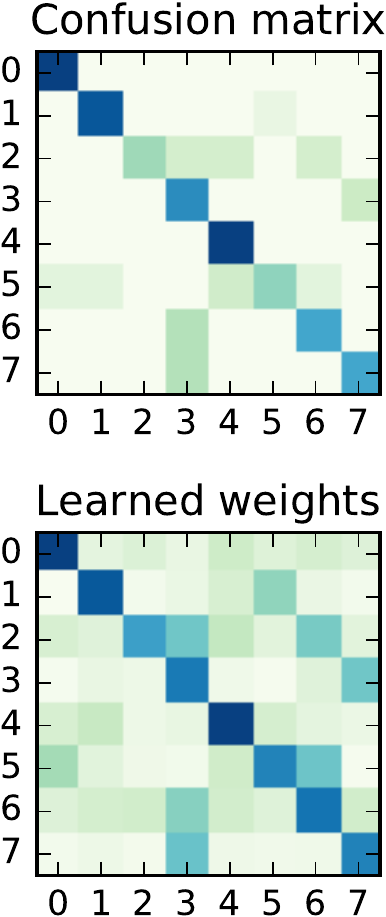}}\hspace{0.28cm}
\subfloat[annot. 20]{\includegraphics[width=0.12\linewidth]{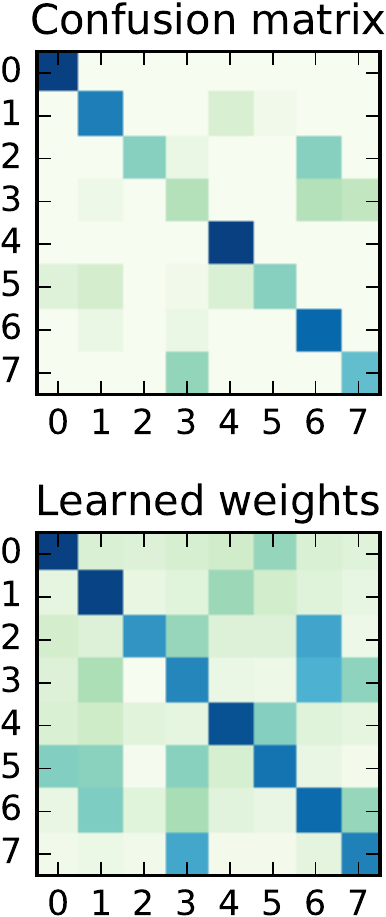}}\hspace{0.28cm}
\subfloat[annot. 23]{\includegraphics[width=0.12\linewidth]{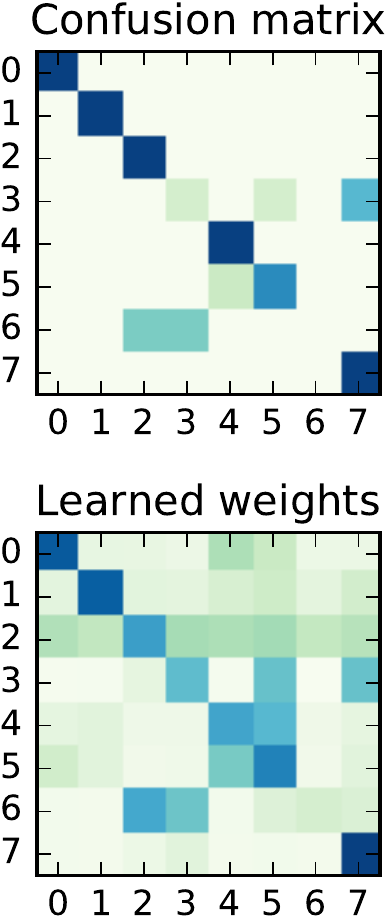}}\hspace{0.28cm}
\subfloat[annot. 39]{\includegraphics[width=0.12\linewidth]{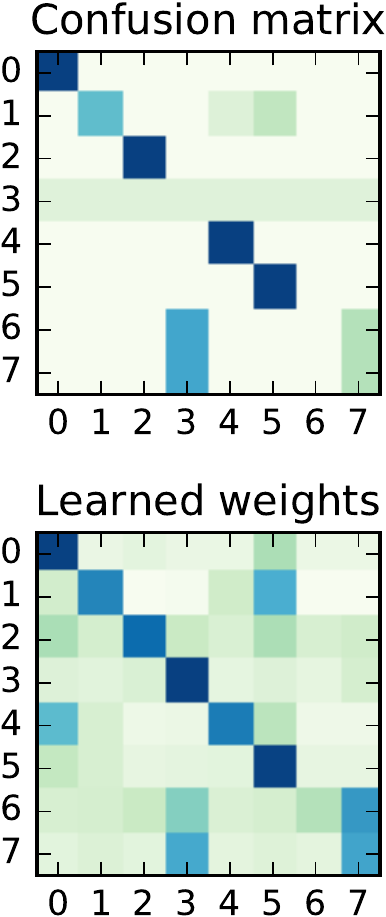}}\hspace{0.28cm}
\subfloat[annot. 45]{\includegraphics[width=0.12\linewidth]{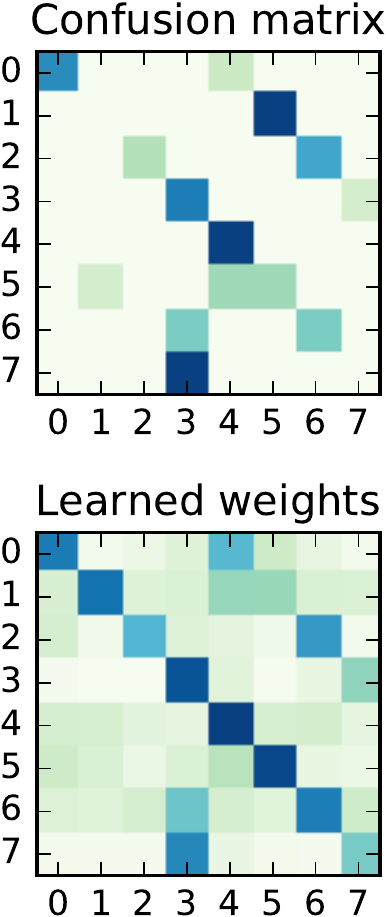}}%
\caption{Comparison between the learned weight matrices $\textbf{W}^r$ and the corresponding true confusion matrices.}
\label{fig:conf_mats_reuters}
\vspace*{-0.3cm}
\end{figure*}

\begin{table*}[!htbp]
\caption{Results for MovieReviews (MTurk) dataset.}
\label{table:results_moviereviews}
\centering
\begin{tabular}{llll}
\toprule
Method & MAE & RMSE & $R^2$\\
\midrule
%sLDA-MEAN \cite{rodrigues2017learning} & - & - & 31.573 ($\pm$ 1.647) \\
MA-sLDAr \cite{rodrigues2017learning} & - & - & 35.553 ($\pm$ 1.282) \\
DL-MEAN & 1.215 ($\pm$ 0.048) & 1.498 ($\pm$ 0.050) & 31.496 ($\pm$ 4.690) \\
DL-EM & 1.201 ($\pm$ 0.046) & 1.482 ($\pm$ 0.048) & 32.974 ($\pm$ 4.457) \\
DL-DN \cite{guan2017said} & 1.270 ($\pm$ 0.021) & 1.549 ($\pm$ 0.022) & 26.775 ($\pm$ 2.102) \\
DL-WDN \cite{guan2017said} & 1.261 ($\pm$ 0.016) & 1.541 ($\pm$ 0.018) & 27.597 ($\pm$ 1.763) \\
\midrule
DL-CL (S) & 1.228 ($\pm$ 0.041) & 1.508 ($\pm$ 0.044) & 30.560 ($\pm$ 4.101) \\
DL-CL (S+B) & 1.163 ($\pm$ 0.031) & 1.440 ($\pm$ 0.033) & 37.086 ($\pm$ 2.407) \\
DL-CL (B) & 1.130 ($\pm$ 0.025) & 1.411 ($\pm$ 0.028) & 39.276 ($\pm$ 2.374) \\
\midrule
DL-TRUE & 1.050 ($\pm$ 0.029) & 1.330 ($\pm$ 0.036) & 45.983 ($\pm$ 2.895) \\
\bottomrule
\end{tabular}
\end{table*}

Of the 25000 images in the Dogs vs Cats dataset, 50\% were used for training and the remaining for testing the different approaches. In order to account for the effect of the random initialization that is used for most of the parameters in the network, we performed 30 executions of all approaches and report their average accuracies in Table~\ref{table:classification_results}. We can immediately verify that both the EM-based and the crowd layer (CL) approaches significantly outperform the majority voting (DL-MV) and Dawid \& Skene (DL-DS) baselines, thus demonstrating the gain of learning from the answers of multiple annotators directly rather than relying on aggregation schemes prior to training. As for the DL-DN and DL-WDN approaches from \cite{guan2017said}, we can observe that, although they also outperform the DL-MV and DL-DS baselines, their accuracy is inferior to that of the proposed DL-CL, which can be explained by the fact that DL-DN and DL-WDN are unable to correct the annotators' biases (\eg confusing class 2 with class 4). Furthermore, it important to recall that due to two-stage procedure of DL-WDN, its computational time can be significantly higher than DL-CL. Regarding the different variants of the proposed crowd layer, we can verify that the MW approach is the one that gives the best average accuracy. In order to better understand what the MW approach is doing, we inspected the weight matrices $\textbf{W}^r$ of each annotator $r$. Figure~\ref{fig:sens_spec_plot} shows the relationship between the diagonal elements of $\textbf{W}^r$ and the true sensitivities and specificities of the corresponding annotators, highlighting the strong linear correlation between the two. This evidences that the proposed crowd layer is able to internally represent the reliabilities of the annotators. 

Having verified that the crowd layer was performing well for simulated annotators, we then moved on to evaluating it in real data from Amazon Mechanical Turk (AMT). For this purpose, we used the image classification dataset from \cite{rodrigues2017learning} adapted from part of the LabelMe data \cite{Russell2008}, whose goal is to classify images according to 8 classes: ``highway", ``inside city", ``tall building", ``street", ``forest", ``coast", ``mountain" or ``open country". It consists of a total of 2688 images, where 1000 of them were used to obtain labels from multiple annotators from Amazon Mechanical Turk. Each image was labeled by an average of 2.547 workers, with a mean accuracy of 69.2\%. The remaining 1688 images were using for evaluating the different approaches. 

Since the training set is rather small, we use the pre-trained CNN layers of the VGG-16 deep neural network \cite{simonyan2014very} and apply only one FC layer (with 128 units and ReLU activations) and one output layer on top, using 50\% dropout. The last column of Table~\ref{table:classification_results} shows the obtained results. We can once more verify that DL-EM, DL-WDN and DL-CL approaches outperform the majority voting and Dawid \& Skene baselines, and also the probabilistic approaches proposed in \cite{rodrigues2017learning} based on supervised latent Dirichlet allocation (sLDA), being the proposed crowd layer (DL-CL) the approach that again gives the best results. However, unlike for the Dogs \mbox{vs.} Cats dataset, the differences between the different function mappings $f_r$ for the crowd layer (CL) become more evident. This can be justified by the ability of the MW version to be able to model the biases of the annotators. Indeed, if we compare the learned weight matrices $\textbf{W}^r$ with the respective true confusion matrices of the annotators, we can notice how they resemble each other. Figure~\ref{fig:conf_mats_reuters} shows this comparison for 6 annotators, where the color intensity of the cells increases with the relative magnitude of the value, thus demonstrating that the crowd layer is able to learn the labeling patterns of the annotators.

\subsection{Text regression}

As previously mentioned, one of the key advantages of the proposed crowd layer is its straightforward extension to other types of target variables. In this section, we consider a regression problem based on the dataset also introduced in \cite{rodrigues2017learning}. This dataset consists of 5006 movie reviews, where the goal is to predict the rating given to the movie (on a scale of 1 to 10) based on the text of the review. Using AMT, the authors collected an average of 4.96 answers from a pool of 137 workers for 1500 movie reviews. The remaining 3506 reviews were used for testing. Letting the (continuous) output of the bottleneck layer be denoted $\mu$, we considered 3 variants of the proposed crowd layer with different annotator-specific functions $f_r$: a per-annotator scale parameter $s^r$, such that $f_r(\mu) = s^r \mu$ (referred to as ``S"); a per-annotator bias parameter $b^r$, such that $f_r(\mu) = \mu + b^r$ (``B"); and a version with both: $f_r(\mu) = s^r \mu + b^r$ (``S+B"). The base neural network architecture used for this problem consists of a 3x3 convolutional layer with 128 features and 5x5 max pooling, a 5x5 convolutional layer with 128 features and 5x5 max pooling, and a FC layer with 32 hidden units. All layers, except for the output one, use ReLU activations. 

The proposed DL-CL is compared with: a neural network trained on the mean answer of the annotators (DL-MEAN) and the approach from \cite{rodrigues2017learning} based on supervised LDA. In order to make the baselines even more competitive, we further propose a new variant of the EM algorithm described earlier that follows the same approach as the extension proposed in \cite{Raykar2010} for regression problems. This approach assumes the following model for the annotators answers given the ground truth: $p(y_n^r|z_n) = \N(y_n^r|z_n, 1/\lambda^r)$. Although the formulation in \cite{Raykar2010} relies on the probabilistic interpretation of the linear regression model to develop an EM algorithm for learning, we can nevertheless adapt the resultant EM algorithm by replacing the linear regression model with a deep neural network. The final iterative procedure then alternates between computing the adjusted ground truth (E-step) and re-estimating the neural network and the annotators' parameters (M-step). Finally, although Guan et al. \shortcite{guan2017said} do not discuss extensions to regression, we also developed variants of DL-DN and DL-WDN for continuous output variables. For the DL-WDN approach, we considered different weighting functions for combining the answers of the multiple annotators, namely: a single weight per annotator, a single bias, or both. We experimented with the different alternatives and found that using a per-annotator bias for combining the answers of the multiple annotators gives the best results. 
 %For the DL-WDN approach, we experimented with different weighting functions among $\hat{y} = \sum_{r=1}^R w^r y^r$ and found that using a per-annotator bias for combining the answers of the multiple annotators gives the best results. 

\begin{figure}[!t]
\centering
%\fbox{\rule[-.5cm]{0cm}{4cm} \rule[-.5cm]{4cm}{0cm}}
\includegraphics[width=0.6\linewidth]{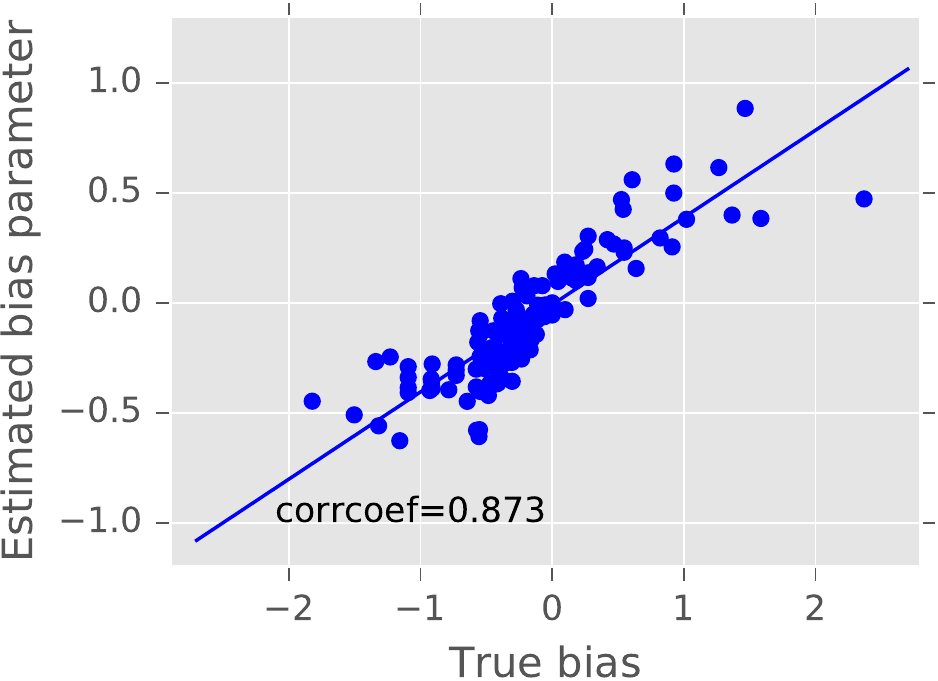}
\caption{Relationship between the learned $b^r$ parameters and the true biases of the annotators.}
\label{fig:bias_plot}
\vspace*{-0.3cm}
\end{figure}

\begin{table*}[!htbp]
\caption{Results for CoNLL-2003 NER (MTurk) dataset.}
\label{table:results_conll}
\centering
\begin{tabular}{llll}
\toprule
Method & Precision & Recall & F1\\
\midrule
%MultiCRF-MAX \cite{Rodrigues2013b} & 0.841 & 0.371 & 0.515 \\
%CRF-MV \cite{Rodrigues2013b} & 0.455 & 0.809 & 0.582 \\
CRF-MA \cite{Rodrigues2013b} & 0.494 & 0.856 & 0.626 \\
%CRF (True) & 0.791 & 0.804 & 0.748 \\
DL-MV & 0.664 ($\pm$ 0.017) & 0.464 ($\pm$ 0.021) & 0.546 ($\pm$ 0.014)  \\
DL-EM & 0.679 ($\pm$ 0.012) & 0.499 ($\pm$ 0.010) & 0.575 ($\pm$ 0.008)  \\
DL-DN \cite{guan2017said} & 0.723 ($\pm$ 0.009) & 0.459 ($\pm$ 0.014) & 0.562 ($\pm$ 0.012)  \\
DL-WDN \cite{guan2017said} & 0.611 ($\pm$ 0.063) & 0.480 ($\pm$ 0.058) & 0.534 ($\pm$ 0.042)  \\
\midrule
DL-CL (VW) & 0.709 ($\pm$ 0.013) & 0.472 ($\pm$ 0.020) & 0.566 ($\pm$ 0.016)  \\
DL-CL (VW+B) & 0.603 ($\pm$ 0.013) & 0.609 ($\pm$ 0.012) & 0.606 ($\pm$ 0.007)  \\
DL-CL (MW) & 0.660 ($\pm$ 0.018) & 0.593 ($\pm$ 0.013) & 0.624 ($\pm$ 0.010)  \\
\midrule
DL-TRUE & 0.711 ($\pm$ 0.013) & 0.740 ($\pm$ 0.009) & 0.725 ($\pm$ 0.008)  \\
\bottomrule
\end{tabular}
\end{table*}

Table~\ref{table:results_moviereviews} shows the obtained results for 30 runs of the different approaches, where we verify that the proposed crowd layer, particularly the ``B" variant, significantly outperforms all the other methods. In order to better understand what the crowd layer in the ``B" variant is doing, we plotted learned $b^r$ values in comparison with the true biases of the annotators, computed as the average difference between their answers and the ground truth. Figure~\ref{fig:bias_plot} shows this comparison, in which we can verify that the learned values of $b^r$ are highly correlated with the true biases of the annotators, thus showing that crowd layer is able to account for annotator bias when learning from the noisy labels of multiple annotators.

\subsection{Named entity recognition}

Lastly, we evaluated the proposed crowd layer on a named entity recognition (NER) task. For this purpose, we used the AMT dataset introduced in \cite{Rodrigues2013b} which is based on the 2003 CoNLL shared NER task \cite{TjongKimSang2003}, where the goal is to identify the named entities in the sentence and classify them as persons, locations, organizations or miscellaneous. The dataset consists of 5985 labeled sentences using a pool of 47 workers. The remaining 3250 sentences of the original dataset were used for testing. 
The neural network architecture used for this problem consists of a layer of 300-dimensional word embeddings initialized with the pre-trained weights of GloVe \cite{pennington2014glove}, followed by a 5x5 convolutional layer with 512 features, whose output is then fed to a GRU cell with a 50-dimensional hidden state. The individual hidden states of the GRU are then passed to a FC layer with a softmax activation. The crowd layer uses the same annotator function mappings $f_r$ used for image classification. 

The proposed crowd layer is compared with same baselines considered for the classification problems. As previously explained, the EM approach is hard to generalize to sequence labelling problems due to marginalization over the latent ground truth sequences in \mbox{Eq.} (\ref{eq:expected_loglikelihood}). In order to make this marginalization tractable, we assume a fully factorized distribution of the posterior approximation $q(\textbf{z}_n)$, such that $q(\textbf{z}_n) = \prod_{t=1}^T q(z_{nt})$, where $T$ denotes the length of the sequence.\footnote{Please note that, while this makes EM tractable, the computational complexity of the E-step is now increased to $\mathcal{O}(N T R)$.} Although the focus of this paper is on deep learning approaches, for the sake of completeness, we also compare with the results of the multi-annotator approach from \cite{Rodrigues2013b} based on conditional random fields (CRF-MA). 
%For the EM approach, in order to make the marginalization over the latent ground truth sequences in Eq. \ref{eq:expected_loglikelihood} tractable, we assumed a fully factorized distribution of the posterior approximation $q(\textbf{z}_n)$, such that $q(\textbf{z}_n) = \prod_{t=1}^T q(z_{nj})$, where $T$ denotes the length of the sequence. 
Table~\ref{table:results_conll} shows the obtained average results, which clearly demonstrate that the proposed approach significantly outperforms all the other methods, and provides similar results to those of CRF-MA, while reducing the training time by at least one order of magnitude when compared to the latter (minutes instead of several hours on a Core i7 with 32GB of RAM and a NVIDIA GTX 1070). 

\section{Conclusion}

This paper proposed the \textit{crowd layer} - a novel neural network layer that enables us to train deep neural networks end-to-end, directly from the labels of multiple annotators and crowds, using backpropagation. Despite its simplicity, the crowd layer is able to capture the reliabilities and biases of the different annotators and adjust the error gradients that are backpropagated during training accordingly. As our empirical evaluation shows, the proposed approach outperforms other approaches that rely on the aggregation of the annotators' answers prior to training, as well as other methods from the state-of-the-art which often rely on more complex, harder to setup and more computationally demanding EM-based approaches. Furthermore, unlike the latter, the crowd layer is trivial to generalize beyond classification problems, which we empirically demonstrate using real data from Amazon Mechanical Turk for text regression and named entity recognition tasks. 

\subsubsection*{Acknowledgments}

The research leading to these results has received funding from the People Programme (Marie Curie Actions) of the European Union's Seventh Framework Programme (FP7/2007-2013) under REA grant agreement no. 609405 (COFUNDPostdocDTU), and from the European Union?s Horizon 2020 research and innovation programme under the Marie Sklodowska-Curie Individual Fellowship H2020-MSCA-IF-2016, ID number 745673.

\bibliographystyle{apalike}

\end{document}